%% file: sample-aamas19.tex
\documentclass[sigconf]{aamas}  

\usepackage{balance}  

\usepackage{booktabs}

\DeclareMathOperator*{\argmax}{argmax}
\usepackage{algorithm}
\usepackage{mathtools}
\usepackage{color}
\usepackage{algorithmic}
\usepackage{graphicx}
\usepackage{subfigure}
\usepackage{float}
\usepackage{floatflt}
\usepackage{wrapfig}
\usepackage{stfloats} 

\setcopyright{ifaamas}  

\acmDOI{}  

\acmISBN{}  

\acmConference[AAMAS'19]{Proc.\@ of the 18th International Conference on Autonomous Agents and Multiagent Systems (AAMAS 2019)}{May 13--17, 2019}{Montreal, Canada}{N.~Agmon, M.~E.~Taylor, E.~Elkind, M.~Veloso (eds.)}  

\acmYear{2019}  

\copyrightyear{2019}  

\acmPrice{}  

\begin{document}

\title{Context-Aware Policy Reuse}  


\author{Siyuan Li}
\affiliation{%
	\institution{IIIS, Tsinghua University}
}
\email{sy-li17@mails.tsinghua.edu.cn}

\author{Fangda Gu}
\affiliation{%
	\institution{Tsinghua University}
}
\email{gfd15@mails.tsinghua.edu.cn}

\author{Guangxiang Zhu}
\affiliation{%
	\institution{IIIS, Tsinghua University}
}
\email{guangxiangzhu@outlook.com}

\author{Chongjie Zhang}
\affiliation{%
	\institution{IIIS, Tsinghua University}
}
\email{chongjie@tsinghua.edu.cn}


\begin{abstract}  
Transfer learning can greatly speed up reinforcement learning for a new task by leveraging policies of relevant tasks.
Existing works of policy reuse either focus on selecting a single best source policy for {reuse} without considering contexts, or fail to guarantee learning  an optimal policy for a target task.
To improve transfer efficiency and guarantee optimality, we develop a novel policy reuse method, called {\em Context-Aware Policy reuSe} (CAPS), that enables multi-policy {reuse}. Our method learns when and which source policy is best for reuse, as well as when to terminate its reuse. CAPS provides theoretical guarantees in convergence and optimality for both source policy selection and target task learning. Empirical results on a grid-based navigation domain and the Pygame Learning Environment demonstrate that CAPS significantly outperforms other state-of-the-art policy reuse methods.
\end{abstract}

%

\keywords{policy reuse; transfer learning; reinforcement learning}  

\maketitle


\input{samplebody-conf}


\bibliographystyle{ACM-Reference-Format}  
\balance  
\bibliography{caps}  

\end{document}

%% file: samplebody-conf.tex
\section{Introduction}

Reinforcement learning (RL) \cite{sutton1998reinforcement} has recently shown
considerable successes of achieving human-level control in challenging tasks \cite{silver2017mastering,mnih2015human}.
However, learning each task independently and from scratch requires vast experiences, and thus is  inefficient for practical problems.
Transfer learning has been actively studied for accelerating RL by making use of prior knowledge \cite{torrey2009transfer}.
Extensive transfer learning research aims to reuse source policies to speed up the learning on a new target task \cite{gupta2017learning,barreto2017successor,parisotto2015actor,fernandez2013learning}.

Many existing policy reuse approaches focus on finding a single best source policy for {reuse}, e.g., by measuring MDP similarity \cite{song2016measuring,ammar2014automated}, through online exploration using multi-armed bandit methods \cite{mazumdar2017multi,li2017optimal}, or via an optimism-under-uncertainty approach \cite{azar2013regret}. However,
such single-policy {reuse} is not efficient enough, because
it is more often that multiple source policies are partially useful for learning a new task.
Although some multi-policy {reuse} methods have been proposed to concurrently utilize multiple source policies in the target task learning, those methods suffer from limitations, e.g., restricting the way of obtaining source policies \cite{brunskill2014pac}, converging to locally optimal termination functions \cite{comanici2010optimal}, or requiring a model of learning environment \cite{mann2014time}.
Our work mainly focuses on the scenarios where source and target tasks have the same state and action space. For the problems of different state and action spaces, a mapping between source and target tasks is needed \cite{fachantidis2015transfer}.

In this paper, we propose a novel model-free multi-policy reuse method, called {\em Context-Aware Policy reuSe} (CAPS).
Our approach learns an optimal source selection policy, which specifies the most appropriate source policy to reuse based on contexts (i.e., a subset of states).
In addition, CAPS provides a convergence guarantee to an optimal target policy, agnostic to the usefulness of the source policies and how to acquire them (which can be either learned from prior experience or provided by advisors).
To improve transfer efficiency and support temporally-extended policy reuse, CAPS utilizes a {\em call-and-return} execution model and concurrently learns when to reuse which source policy, as well as when to terminate its reuse.
Moreover, CAPS augments the source policy library with primitive policies to ensure the completeness of the action space.
Our approach also exploits the overlapping between source policies and enables concurrent Q-value updates for multiple source policies in order to make full use of experiences.
{ CAPS assumes that the action space of the learning problem is discrete and the state-action space is partially shared between source and target tasks.}
Empirical results in a grid-based navigation domain as well as the Pygame Learning Environment (PLE) \cite{tasfi2016PLE} show that CAPS (i) learns the optimal source selection policy with temporally-extended policy reuse and speeds up the target task learning significantly even when its transition function is different from those of source tasks; (ii) outperforms state-of-the-art transfer algorithms remarkably when multiple source policies are useful; and (iii) achieves the same performance as, if not better than, single-policy reuse methods in situations where only one source policy is useful.

This paper is structured as follows. Section 2 summarizes threefold related work. In Section 3, we describe the background knowledge and problem statement. Section 4 firstly presents a new take on multi-policy reuse, and then provides our approach of learning source selection policy and termination functions for policy reuse, followed by the theoretical results. Section 5 presents an empirical illustration of both toy and deep-learning experiments. Finally, Section 6 concludes and discusses avenues for future work.

\section{Related Work}
{\bf Policy Reuse.} Most state-of-the-art policy reuse methods lack theoretical guarantees and analysis.
\cite{rosman2016bayesian} proposes a policy reuse method, mainly working on short-lived sequential policy selection without learning a full policy.
On the contrary, our method converges to the optimal policy for a target task. To reuse a selected source policy,
\mbox{\cite{fernandez2013learning,li2017optimal}} combine it with a random policy according to the episode length. Such an ad hoc reuse strategy has a great effect on the transfer performance and its hyperparameters are hard to tune. In contrast, CAPS automatically learns when and which source policy to reuse.
\cite{laroche2017transfer} reuses transition samples obtained from one task to accelerate learning of another, but it is constrained with the assumption that transition samples can be shared across tasks.
\cite{ammar2015unsupervised} initializes a target policy with a single mapped source policy via unsupervised manifold alignment, which is unable to reuse multiple source policies.
\cite{wang2018target,brys2014combining} focus on value-based reuse, which reuse the value functions of previous policies. Our approach assumes no prior knowledge about the representations of the source policies and still works when the value functions for source policies are not available.

{\bf Multi-Task Learning (MTL)} co-learns a set of tasks jointly via some shared knowledge \cite{caruana1998multitask,fachantidis2015transfer,brunskill2013sample},
so an agent needs the environment information of all the tasks.
However, our approach does not require to know source task models.
MLSH is a hierarchical MTL method learning a master policy and several sub-policies with fixed length \cite{frans2017meta}. However, unlike our method, it cannot learn when to terminate the sub-policies autonomously. Moreover, MLSH has no theoretical guarantee for optimal convergence.
UVFA learns the policies for multiple goals in the same environment simultaneously with the goal information \cite{schaul2015universal}, but our method requires no prior knowledge about goals of the source and target tasks.

{\bf Option Learning.}
In contrast to the works of option discovery \cite{bacon2017option,harutyunyan2017learning,harb2017waiting,jong2008utility}, CAPS focuses on multi-policy reuse.
Although our approach shares some similarity in the termination function learning as \cite{bacon2017option}, CAPS seamlessly integrates with policy reuse and provides the convergence and optimality guarantee of learning the target policy, which \cite{bacon2017option} cannot provide.
The objectives of CAPS and Option-Critic (OC) \cite{bacon2017option} are orthogonal. While OC  learns multiple source policies in the form of options from scratch, CAPS transfers the learned policies efficiently to a new task.
Some methods have been proposed to reuse options, but they suffer from several limitations.
For example, \cite{sutton1999between,tessler2017deep} assumes the given options are fixed and their reuse cannot be adapted to a target task structure for more efficient transfer.
Source policies in \cite{brunskill2014pac} are restricted to be learned in a PAC-learning way to obtain a $\epsilon$-optimal option library.
\cite{mann2014time} learns terminations for policies via value iteration with an environment model.
\cite{comanici2010optimal} assumes the given policy library is sufficient and converges to a locally optimal termination function.
However, CAPS has no requirement to an environment model, sufficiency of the source policy library or how it is acquired.
In addition, CAPS adaptively learns terminations for source policies.

\section{Preliminaries and Problem Statement}
This paper focuses on RL tasks, whose environments can be modeled by Markov Decision Processes (MDP).
An MDP consists of a set of states $\mathcal S$, a set of actions $\mathcal A$, a transition function $P$, and a reward function $\mathcal R$.
At each time step, an agent chooses and executes an action $a$ on the current state $s$, and then receives a reward $R(s, a)$ and observes the next state $s'$ according to transition function $P(s'|s,a)$. A policy $\pi: \mathcal S \to \mathcal A$ specifies an action for each state and its state-action value function $Q_{\pi}(s, a) = \mathbb{E}_\pi[\sum_{t=0}^\infty \gamma^t r_{t+1} | s_0 = s,  a_0 = a]$ is the expected return for executing action $a$ on state $s$ and following policy $\pi$ afterwards. $\gamma \in [0,1)$ is a discount factor.
A greedy policy $\pi$ with respect to a value function $Q$ is given by $\pi(s)=\argmax_{a \in \mathcal A} Q(s, a)$ for all state $s \in \mathcal S$.
The goal of an RL algorithm on an MDP is to find an optimal policy that maximizes the expected return. Q-learning learns the optimal Q function, which yields an optimal policy \cite{watkins1992technical}.

{\bf A Policy Reuse Problem} is defined as following: given a set of source policies $\Pi_s=\{\pi_1,\pi_2,...,\pi_n\}$ and a new target task $g$, the goal is to quickly learn an optimal policy for the target task by exploiting knowledge from source policies.
{This formulation is a standard online policy reuse framework that is also used in \cite{fernandez2013learning} and \cite{li2017optimal}.}
Source policies can be either learned for different but relevant tasks or heuristically designed by humans. In this paper, we assume that each source policy and the target task have the same state-action space. This assumption can be relaxed
if the states and actions in the source tasks could be mapped to a target task.
The policy reuse problem has two related objectives: finding the optimal policy $\pi_g^*$ for the target task and learning the optimal source selection policy $\pi_{\Pi_s}^*: \mathcal S \to \Pi_s$ for reusing source policies during the learning.
The optimal source selection policy $\pi_{\Pi_s}^*$ should be consistent with $\pi_g^*$, that is, if source policy $\pi_i \in \Pi_s$ is selected by $\pi_{\Pi_s}^*$ for state $s$, then it should select the same action on state $s$ as the optimal target policy, i.e., $\pi_i(s) = \pi_g^*(s)$.

\section{Approach}
We aim to enable an agent to quickly learn an optimal policy for a new target task by leveraging knowledge from multiple source policies. Selecting a single policy to reuse is not efficient when provided with a source policy library where multiple policies are partly useful. Therefore, it is essential to identify both when (i.e., on which states) and which source policy is the most appropriate to reuse.

We develop a policy reuse method, called {\em Context-Aware Policy reuSe} (CAPS).
 By exploiting the option framework \cite{sutton1999between}, CAPS formulates source policy selection as an inter-option learning problem, whose solution is called a {\em source selection policy} specifying the choice of a source policy to reuse for each state.
In the formulation, the source policy library is expanded with primitive policies to ensure the optimality of the learned target policy no matter whether the usefulness of source policies is sufficient or not.
To improve transfer efficiency and support temporally-extended policy reuse, CAPS uses the {\em call-and-return} model for reusing selected policies, where the execution of  a selected policy is returned until completion according to its termination function \cite{precup1998theoretical}.
{Once the best option is selected, an agent may take the best actions for multiple steps, instead of making an option choice for every state.}
CAPS simultaneously learns the source selection policy and the termination function for each source policy and primitive policy.
Theoretical guarantees in convergence and optimality are provided for CAPS in learning both the source selection policy and the target task policy.

In the rest of this section, we will first describe our formulation of multi-policy reuse as inter-option learning, then present the CAPS learning algorithm, and finally analyze the theoretical guarantees.

\subsection{Formulation as Inter-Option Learning}

We formulate multi-policy reuse as an inter-option learning problem. Options are temporally-abstracted policies for taking actions over a period of time \cite{sutton1999between}.
An option $o \in \mathcal O$ is defined by a triple $(\pi_o, \mathcal I, \beta_o)$, where $\pi_o$ is an intra-option policy, $\mathcal I \subseteq \mathcal S$ is an initiation state set, and $\beta_o:  \mathcal S \to  [0,1] $ is a termination function that specifies the probability of option $o$ terminating on each state $s \in \mathcal S$.  Any MDP endowed with a set of options becomes a semi-MDP, which has a corresponding optimal option-value function $Q_{\mathcal O}(s, o)$ over options.

In our formulation, we create a set of source options $\mathcal O_s$ from the given source policy library $\Pi_s$. For each source policy $\pi \in \Pi_s$, we instantiate an option $o = (\pi_o, \mathcal I, \beta_{\theta_o})$, where its intra-option policy $\pi_o = \pi$, its initiation set $\mathcal I$ is the whole state space, and its termination function $\beta_{\theta_o}$ is defined by a sigmoid function with a differentiable parameter $\theta_o$\footnote{$\theta_o$ is overloaded and represents a function of state $s$ and option $o$, which is parameterized by $\theta_o$.}:
\begin{align*}
\beta_{\theta_{o}}(s)= \frac{1}{1+e^{-\theta_{o}(s)}}.
\end{align*}

With this formulation, an inter-option policy corresponds to a source selection policy. Reusing a selected policy is applying its action selection on the target task. However, such policy reuse will lead to a suboptimal policy for the target task when
the source policy library is not sufficient (i.e.,
the actions of an optimal target policy for all states are not identical to any actions of source policies.).
To enable the optimality guarantee for learning the target task, we augment the source policy library $\Pi_s$ with primitive policies $\Pi_p=\{\pi_{1},\pi_2,...,\pi_{\left | A \right | }\}$, where policy $\pi_{i} \in \Pi_p$ takes action $a_i \in A$ for all states. Correspondingly, the source option set $\mathcal O_s$ is expanded to an option set $\mathcal O$ by including primitive options $\mathcal O_p$ created from primitive policies $\Pi_p$. Such augmentation ensures that all actions are available to all states{, which enables the optimal guarantees of our approach.
	A random policy cannot replace primitive policies, because it cannot be part of an optimal deterministic policy, which exists for all MDPs.}

To obtain an optimal source selection policy, we need to learn the option-value function $Q_{\mathcal O}^{\pi_{\mathcal O}}(s,o)$, which is defined as the expected discounted future reward starting in $s \in \mathcal I$, taking $o$, and henceforth following an inter-option policy $\pi_{\mathcal O}: \mathcal S \to \mathcal O$,
\begin{align*}
Q_{\mathcal O}^{\pi_{\mathcal O}}(s, o)=E\left \{r_{t+1}+\gamma r_{t+2}+\ldots|s_t=s, o_t=o, \pi_{\mathcal O} \right\}.
\end{align*}
The optimal option value is defined as $Q^*_{\mathcal O}(s,o)=\max_{\pi_{\mathcal O}}Q_{\mathcal O}^{\pi_{\mathcal O}}(s,o)$. As we use the {\em call-and-return}  model of option execution , the option-value function $Q_{\mathcal O}^{\pi_{\mathcal O}}$ also depends on when the execution of selected options terminates. Therefore, in addition to learning the optimal inter-option policy, we need to learn the termination functions for all options as well.

\subsection{Context-Aware Policy Reuse}
\begin{algorithm}[htbp]
	\caption{Context-Aware Policy Reuse}
	\begin{algorithmic}[1]
		\STATE{Instantiate options $\mathcal O$ from source and primitive policies}
		\STATE{Initialize option-value function $Q_{\mathcal O}$ }
		\STATE{Initialize termination function $\beta_{\theta_o}$ for all option $o \in \mathcal O$}
		\FOR{episode$=1..M$}
		\STATE {$s \leftarrow$ initial state }
		\STATE{$o\leftarrow\epsilon\text{-greedy}(Q_{\mathcal O}, \epsilon, \mathcal O,s)$}
		\WHILE{$s$ is not terminal}
		\STATE {Execute $a=\pi_{o}(s)$ and obtain next state $s'$ and reward $r$}	
		\STATE{$Q_{\mathcal O }\leftarrow$Update$Q_{ \mathcal O}$ ($Q_{\mathcal O}, \beta_\theta, \mathcal O, s, a, s', r$)}
		\STATE{Update termination function $\beta_{\theta_o}$ with $Q_{\mathcal O}$ }
		\IF{Option $o$ terminates according to $\beta_{\theta_o}(s')$}
		\STATE{$o\leftarrow\epsilon\text{-greedy}(Q_{\mathcal O}, \epsilon, \mathcal O, s')$}
		\ENDIF
		\STATE{$s \leftarrow s'$}
		\ENDWHILE
		\ENDFOR
	\end{algorithmic}
\end{algorithm}

Given the inter-option learning formulation above, we here present our algorithm for learning both the optimal source selection policy and the termination functions for source policies and primitive policies during temporally-extended policy reuse.

Algorithm 1 illustrates CAPS using a tabular action-value function representation, which is also applicable with a function approximation.
First, a set of policies $\mathcal O$ with parameterized termination functions are created based on the given source policy library $\Pi_s$  and primitive policies (Line 1), as described in the previous subsection.
Then we learn option-value function $Q_{\mathcal O}$ in the {\em call-and-return} model of option execution, where an option $o$ is executed until it terminates based on its termination function $\beta_{\theta_o}$ and then a next option is selected by a policy over options $\pi_{\mathcal O}$, which is $\epsilon$-greedy to $Q_{\mathcal O}$.

\subsubsection{\bf Learning Source Selection Policy}
	\mbox{ }\par
	Algorithm 2 is used to learn option-value function $Q_\mathcal O$, which
	satisfies the Bellman equation analogously to one-step intra-option Q-learning \cite{sutton1998intra}.Since options are temporal abstractions, the expected return of next state $U^*(s',o)$ is proportional to $\beta(s')$, the probability that option $o$ terminates in \mbox{state $s'$.}
\begin{align*}
{\small U^*(s',o)=(1-\beta(s'))Q^*_{\mathcal O}(s',o)+\beta(s')\max_{o' \in \mathcal O}Q^*_{\mathcal O}(s',o').}
\end{align*}

\begin{algorithm}[htbp]
	\caption{Update$Q_{ \mathcal O}$ ($Q_{\mathcal O},\beta_\theta,\mathcal O,s,a,s',r$)}
	\begin{algorithmic}[1]
		\FOR{$o_i \in \mathcal O$}
		\IF{$a=\pi_{o_i}(s)$}
		\STATE	$Q_{\mathcal O}(s,o_i)\leftarrow(1-\alpha)Q_{\mathcal O}(s,o_i)+\alpha(r+\gamma U(s',o_i))$
		\begin{eqnarray*}
			\hspace{12mm}
			{ U(s',o_i)=(1\!\!-\!\!\beta_{\theta_{o_i}}(s'))Q_{\mathcal O}(s',o_i)\!\!+\!\!\beta_{\theta_{o_i}}(s')\max_{o' \in \mathcal O}Q_{\mathcal O}(s',o')}
		\end{eqnarray*}		
		\ENDIF	
		\ENDFOR
		\STATE {\bfseries return } $Q_{\mathcal O}$
	\end{algorithmic}
\end{algorithm}

The value function $U(s',o)$ is an estimate of $U^*(s',o)$.
We update option-value functions for all the options which select the same action as the current action $a$ in  order to make full use of experiences.
Since $\beta_{\theta_{o_i}}(s')$ and $Q_{\mathcal O}(s',o_i)$ are different for each $o_i$ satisfying the condition, $Q_{\mathcal O}(s,o_i)$ is updated differently for those options.

CAPS chooses a proper option $o$ by utilizing $\epsilon$-greedy strategy according to $Q_{\mathcal O}$ (Line 6, 12).
With a probability of $\epsilon$, we randomly choose an option, and with a probability of $1-\epsilon$, we choose the option with the maximum Q-value.
As $\epsilon$ never equals 0, all state-option pairs will be visited infinitely often. 

\subsubsection{\bf Learning Termination Functions for Policy Reuse}
 As the selected source policy cannot be all the same as an optimal target policy, CAPS also needs to learn when to terminate the selected policy.
{Learning termination functions supports temporally-extended policy reuse.}
CAPS learns termination functions in a similar way to \cite{bacon2017option}, aiming to solve a multi-policy transfer learning problem, instead of option discovery. 
The objective of learning termination functions is to maximize the expected return $U$, so we can update the parameters of the termination functions with the following gradient:
\begin{equation}
\frac{\partial U(s_1,o_0)}{\partial\theta_{o_0}}=-\sum_{s',o}\mu_{\mathcal O}(s',o|s_1,o_0)\frac{\partial\beta_{\theta_{o}}(s')}{\partial\theta_o}A_{\mathcal O}(s', o),
\end{equation}
where
\begin{equation*}
\mu_{\mathcal O}(s',o|s_1,o_0)=\sum _{t=0}^{\infty}\gamma ^tP(s_{t+1}=s',o_t=o|s_1,o_0).
\end{equation*}
$P(s_{t+1}\!=\!s',o_t\!=\!o|s_1,o_0)$ is the transition probability from initial condition $(s_1, o_0)$ to $(s',o)$ in $t$ steps.
Advantage function $A_{\mathcal O}(s',o)$ is defined as $Q_{\mathcal O}(s',o)-\max_{o'\in \mathcal O}Q_{\mathcal O}(s',o').${ The reason that we do not use the conventional definition of the advantage function, $Q_{\mathcal O}(s',o)-E_{o'\sim\pi(s')}[Q(s',o')]$, is to ensure the termination functions of the non-optimal options converge to 1, which guarantees the optimality of the learned policy, as shown in the proof of Theorem 2.}

The transition probability in Equation (1) is estimated from samples of the stationary on-policy distribution.
For data efficiency, the discounted factor $\gamma$ is neglected.
So we update the  parameter $\theta_o$ of termination function as follows:
\begin{equation}\label{update_function}
\theta_o\leftarrow\theta_o\!-\!\alpha_{\beta}\frac{\partial\beta_{\theta_o}(s')}{\partial\theta_o}(Q_{\mathcal O}(s',o)-\max_{o'\in O}Q_{\mathcal O}(s',o'))
\end{equation}
to identify transfer contexts (Line 10).

If Q-value of the current option $o$ is not the largest among all the options, its termination probability grows, so the agent has a higher probability to switch to other better source policies.
The termination probability of non-optimal options will converge to 1 eventually.
CAPS achieves  context identification autonomously by learning termination functions of the formulated options.

\subsection{Theoretical Analysis}
In this section, we provide theoretical analysis for CAPS.
As guaranteed by Theorem 1, CAPS learns an optimal source selection policy that chooses the best source policy or primitive policy for each state.
\newtheorem*{theorem3}{Theorem 1 Optimality on Source Selection Policy}
\begin{theorem3}
	\mbox{ }\par
	Given any source policy library $\Pi_s$, bounded rewards $|r_n|\leq R$, learning rates $0\leq \alpha_t \le 1$, and $\sum_{i=1}^{\infty}\alpha_{t^i}(s,a)=\infty$, $\sum_{i=1}^{\infty}\alpha_{t^i}^2(s,a)<\infty$
	for all $(s,a)\in\mathcal S \times \mathcal A$, then
	the CAPS algorithm converges w.p.1 to an optimal source selection  policy.
\end{theorem3}
\begin{proof}
	We apply the update rule of $Q_{\mathcal O}$ to each option $o$ that takes the same action with the current action $a$ taken by an agent:
	\begin{align}
	\label{Qo_update}
	Q_{\mathcal O}(s,o)\leftarrow(1-\alpha)Q_{\mathcal O}(s,o)+\alpha(r+\gamma U(s',o)).
	\end{align}
	Then we subtract $Q^*_{\mathcal O}(s,o)$ from both sides of the update function (\ref{Qo_update}) and defining $\Delta_t(s,o)=Q_{\mathcal O, t}(s,o)-Q^*(s,o)$ together with
	\begin{eqnarray*}
		F_t(s,o)=r +\gamma U(s',o)-Q^*_{\mathcal O}(s,o).
	\end{eqnarray*}
	The learning rule of $Q_{\mathcal O}$ can be seen as the iterative process of Theorem 1 in \cite{jaakkola1994convergence}.
	\begin{small}
		\begin{align*}
		\small
		&\left|E\left\{F_t\left(s,o\right)\right\} \right|= \left| r+\gamma\sum_{s'}P(s'|s,a)U(s',o)-Q^*_{\mathcal O}(s,o)\right| \nonumber \\
		=& \left| r\!+\!\gamma\sum_{s'}P(s'|s,a)U(s',o)\!-\!  (r\!+\!\gamma\sum_{s'}P(s'|s,a)U^*(s',o))\right| \nonumber\\
		=& \Bigg | \gamma\sum_{s'}P(s'|s,a)\ \bigg[ (1-\beta_o(s'))(Q_{\mathcal O}(s',o)-Q^*_{\mathcal O}(s',o))\nonumber\\
		+&\beta_o(s')(\max\nolimits_{o' \in \mathcal O}Q_{\mathcal O}(s', o')-\max \nolimits_{o' \in \mathcal O}Q^*_{\mathcal O}(s', o'))\bigg ]  \Bigg|\\
		\leq &\gamma\sum_{s'}P(s'|s,a)\max\nolimits_{s'',o''}\left|Q_{\mathcal O}(s'',o'')-Q^*_{\mathcal O}(s'',o'')\right|\\
		= & \gamma \max\nolimits_{s'',o''}\left|Q_{\mathcal O}(s'',o'')-Q_{\mathcal O}^*(s'',o'')\right|
		\end{align*}
	\end{small}
	As a result, $E\{F_t(s,o)\}$ has a contraction property.
	\begin{align*}
	&var[F_t(s,o)| \mathcal F_t]=var[r+\gamma U(s',o)|\mathcal F_t],
	\end{align*}
	where $\mathcal F_t=\{\Delta_t,\Delta_{t-1},...,F_{t-1},...\alpha_{t-1},...,1-\alpha_{t-1},...\}$ represents the past at step $t$. Because $r$ is bounded,verifies
	\begin{align*}
	var[F_t(s,o)|\mathcal F_t]\leq C(1+||\Delta_t||^2_W),
	\end{align*}
	where $C$ is some constant and $||\cdot||_{W}$ denotes some weighted maximum norm. Since $\sum_t\alpha_t=\infty,\sum_t\alpha_t^2<\infty$ and $\epsilon$ in Algorithm 2 never equals 0, all the conditions of Theorem 1 in \cite{jaakkola1994convergence} are satisfied.
	$Q_{\mathcal O}$ converges w.p.1 to the optimal Q-function.
	Following a greedy strategy to $Q_{\mathcal O}$
	($o(s)=\argmax_{o \in \mathcal O} Q_{\mathcal O}(s, o)$)
	, CAPS converges to  an optimal source selection policy for policy library $\Pi$.
\end{proof}
Although our basic proof structure of Theorem 1 is based on that of Q-learning, it extends the convergence and optimality guarantees of Q-learning to a more general setting of temporally-extended inter-option learning.
In addition, CAPS is able to converge to an optimal policy for a target task no matter what source policies are given. This theoretical guarantee is provided in Theorem 2.
\newtheorem*{theorem1}{Theorem 2 Optimality on Target Task Learning}
\begin{theorem1}
	\mbox{ }\par
	Given any source policy $\Pi_s$, bounded rewards $|r_n|\leq R$, learning rates $0\leq \alpha_t \le 1$, and $\sum_{i=1}^{\infty}\alpha_{t^i}(s,a)=\infty$, $\sum_{i=1}^{\infty}\alpha_{t^i}^2(s,a)<\infty$
	for all $(s,a)\in\mathcal S \times \mathcal A$, then
	the CAPS algorithm converges w.p.1 to an optimal policy $\pi_g^*$ for any target task $g$.
	
\end{theorem1}

\begin{proof}
	The termination function of $\forall o \in \mathcal O$ is defined
	as:
	\begin{align*}
	\beta_{\theta_{o}}(s)= \frac{1}{1+e^{-\theta_{o}(s)}}.
	\end{align*}
	The update rule of $\theta$ is:
	\begin{small}
		\begin{align*}
		&\theta_{o}^{t+1}=\theta_{o}^{t}-\alpha_{\beta}\frac{\partial\beta_{\theta_o^{t}}(s)}{\partial\theta_o^t}(Q_{\mathcal O}(s,o)-\max_{o'}Q_{\mathcal O}(s,o'))\\
		&=\theta_{o}^{t}-\alpha_{\beta}\beta_{\theta_o^{t}}(s)(1-\beta_{\theta_o^{t}}(s))(Q_{\mathcal O}(s,o)-\max_{o'}Q_{\mathcal O}(s,o'))
		\end{align*}
	\end{small}
	Let $\theta_{o}^{t+1}=f(\theta_{o}^{t})$ and $o_g(s)=\argmax_{o_i \in \mathcal O} Q_{\mathcal O}(s,o_i)$.
	For arbitrary non-optimal options, i.e., $\forall o_i \in \mathcal O \backslash\{o_g\}$, $f(\theta_{o_i}^{t})$ monotonically increases and $\theta_{o_i}^{t+1}>\theta_{o_i}^{t}$.
	If all state-option pairs can be visited infinitely often, for any non-optimal option $o_i$,
	\begin{align*}
	\lim_{ t\rightarrow \infty}\theta_{o_i}^{t}(s)\rightarrow \infty,
	\end{align*}
	\begin{align*}
	\lim\limits_{ t\rightarrow \infty}\beta_{\theta^{t}_{o_i}}(s)=\frac{1}{1+e^{-\lim\limits_{t\rightarrow \infty}\theta^t_{o_i}(s)}}=1.
	\end{align*}
	According to Theorem 1,
	$Q_{\mathcal O}$ converges to $Q^*_{\mathcal O}$. Because the termination functions of the non-optimal options converge to 1, the greedy policy obtained in
	the call-and-return option execution model is an optimal policy for task $g$.
\end{proof}
Theorem 2 also extends the convergence and optimality guarantees of traditional RL to a  more general learning setting with or without reusing source policies.
{When provided with some source policies, learning with our method
	can exploit those source knowledge and
	significantly speed up the learning of the optimal policy for an unknown target task, which is demonstrated in the following empirical results.}
\vspace{-5pt}
\begin{figure}[htbp]
	\centering
	{	\includegraphics[width=0.47\textwidth]{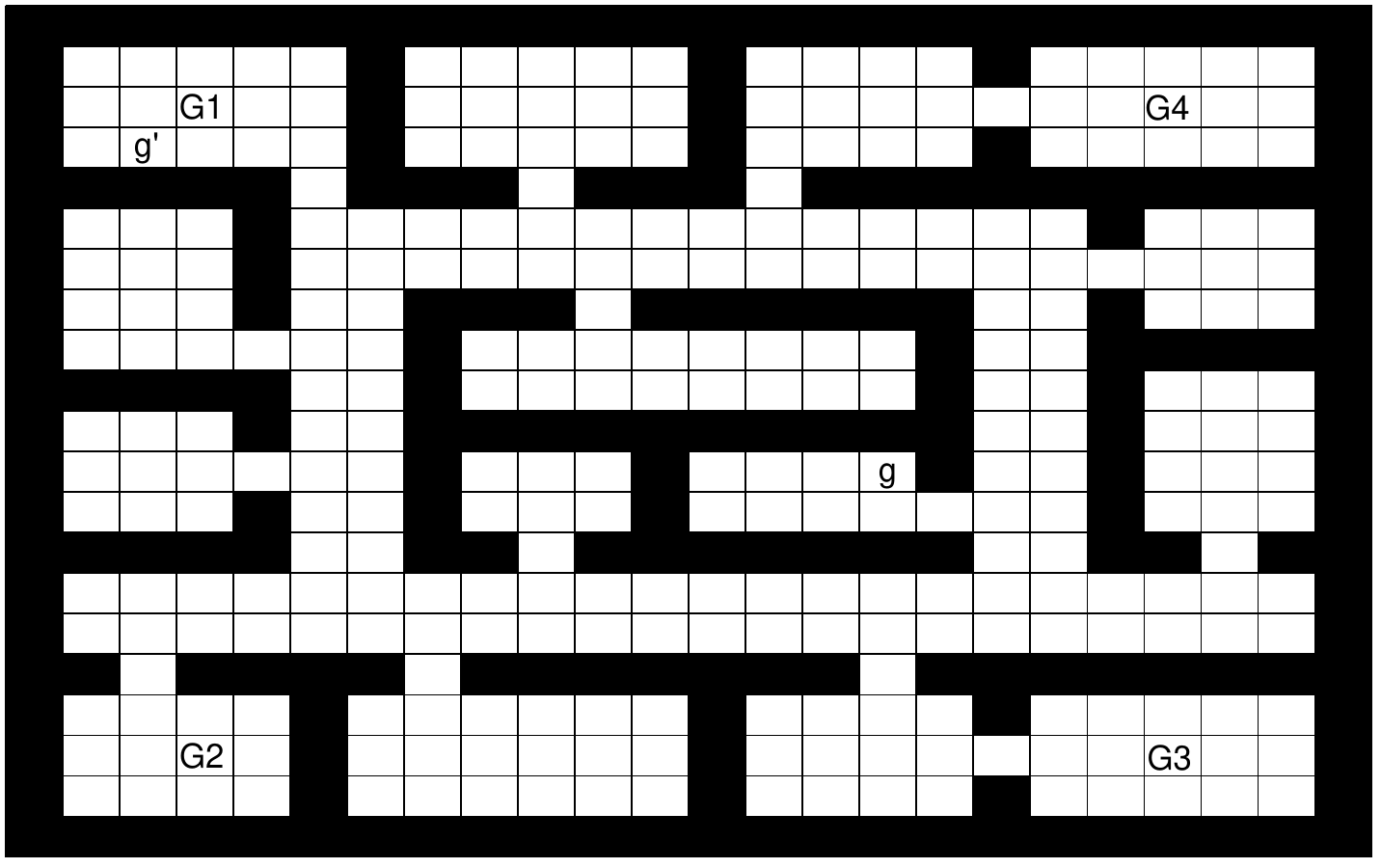}}
	\setlength{\abovecaptionskip}{-8pt}
	\setlength{\belowcaptionskip}{-18pt}
	\caption{Source and target tasks in the map.}
\end{figure}

\section{Empirical Results}
We compare CAPS with state-of-the-art policy reuse algorithms, PRQL \cite{fernandez2013learning} and OPS-TL \cite{li2017optimal}, in addition to Q-learning and CAPS with fixed termination functions. 
We also include a baseline from the option learning literature, Option-Critic (OC) \cite{bacon2017option}.
We first consider a grid-based navigation domain used by \cite{fernandez2013learning} and \cite{li2017optimal}. We then augment all approaches with deep neural networks for function approximation and evaluate them in Pygame Learning Environment (PLE) \cite{tasfi2016PLE}.
In addition, we also do experiments in situations where transition functions of source and target tasks are different.

\subsection{Grid-based Navigation Domain}

In the grid-based navigation domain,
we define states of an agent by grids. Figure 1 shows goals of source and target tasks in the map.
Initial states are randomly set and $G1,G2,G3,G4$ denote goals of source tasks. $g$ and $g'$ represent goals of different target tasks. Action space consists of {\em up, down, left} and {\em right}, four actions, which make an agent move in the corresponding direction with a step size of 1.
The position of an agent is added a noise after each action to make stochastic MDP environment.

Each learning process has been executed 10 times and the maximum episode length $H$ is set as 100.
The learning rates are set to 0.5 for $Q_O$ and 0.2 for termination. 
$\gamma$ is set as 0.95.
Q-learning is executed using an $\epsilon$-$greedy$ strategy with $\epsilon = 1-k/(k+800)$, where $k$ is the episode number.
{{
To compare with OC, we first train the source tasks using the OC model with only one intra-option policy until convergence, so the intra-option policies can be regarded as source policies in our setting. Then we train  a new OC model to learn a target task. The new OC model has 4 intra-option policies, which are initialized with the intra-option policies learned in the source tasks
\footnote{The details of the Option-Critic implementation  is in the Appendix.}.}}

\vspace{-8pt}
\begin{figure}[htbp]
	\centering
	{
		\includegraphics[width=0.47\textwidth]{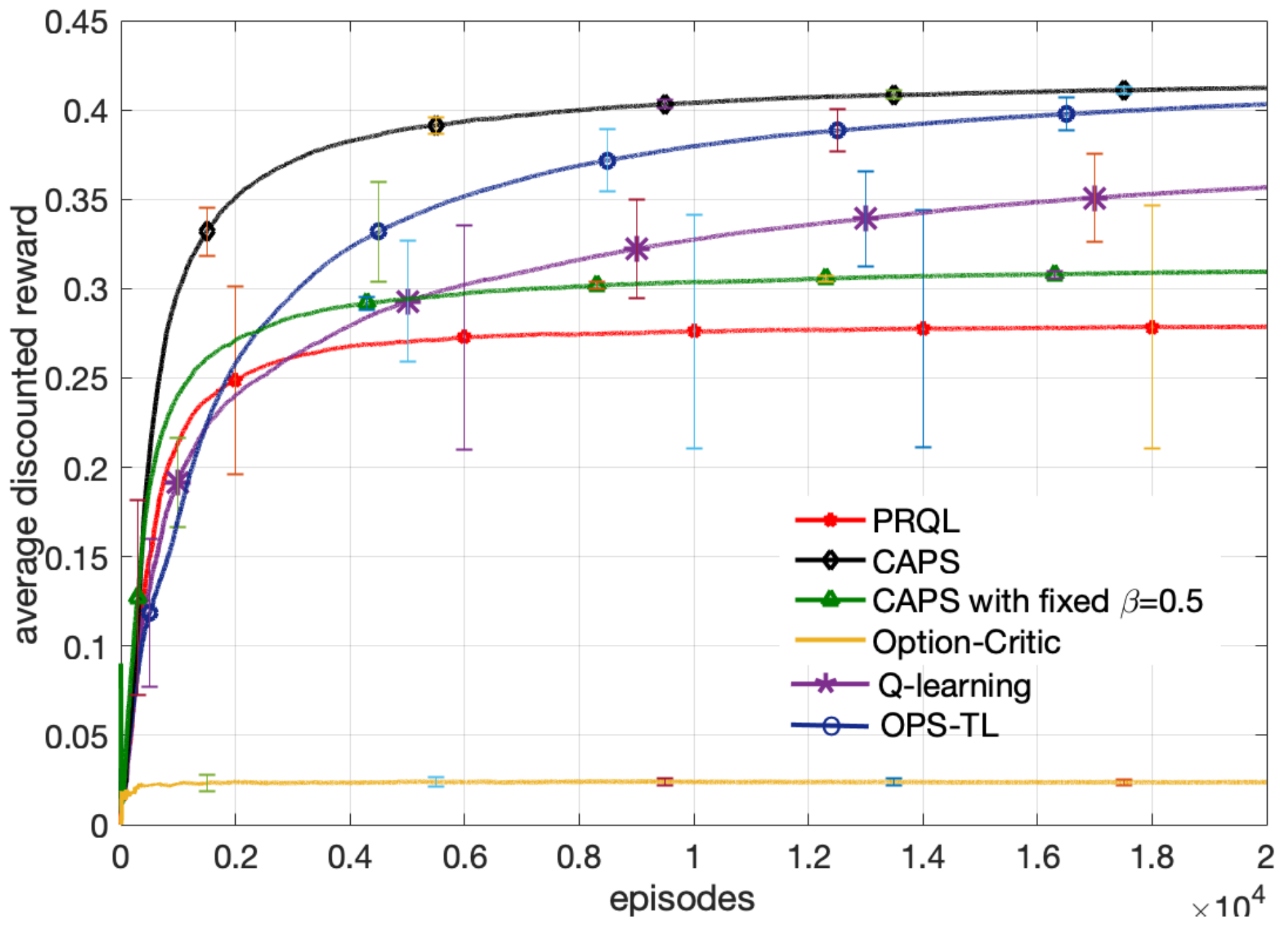}}	
	\setlength{\abovecaptionskip}{-8pt}
	\setlength{\belowcaptionskip}{-9pt}
	\caption{Average discounted rewards of CAPS, PRQL, OPS-TL, OC and Q-learning on target task $g$. }
\end{figure}

Figure 2 shows the learning curves for the target task $g$, where all source tasks are not quite similar to the target task.
CAPS significantly accelerates  the learning process and dramatically outperforms OPS-TL, PRQL and Q-learning.
Furthermore, PRQL results in negative transfers compared to Q-learning in this circumstance.
With termination probability fixed as 0.5, the initial value for $\beta$, CAPS converges to a suboptimal policy, so it is necessary to learn
when to terminate reusing the selected policies.
{{
OC performs dramatically worse than Q-learning, illustrated in Figure 2. The reasons are two-fold. First, the source policies are nearly deterministic, and the action space for the inter-option policy in OC is incomplete. Second,  the co-adaptation of inter-option policy and options for OC could be problematic for reusing learned options.

}}

\begin{figure}[hbpt]
	\centering
	{	\includegraphics[width=0.47\textwidth]{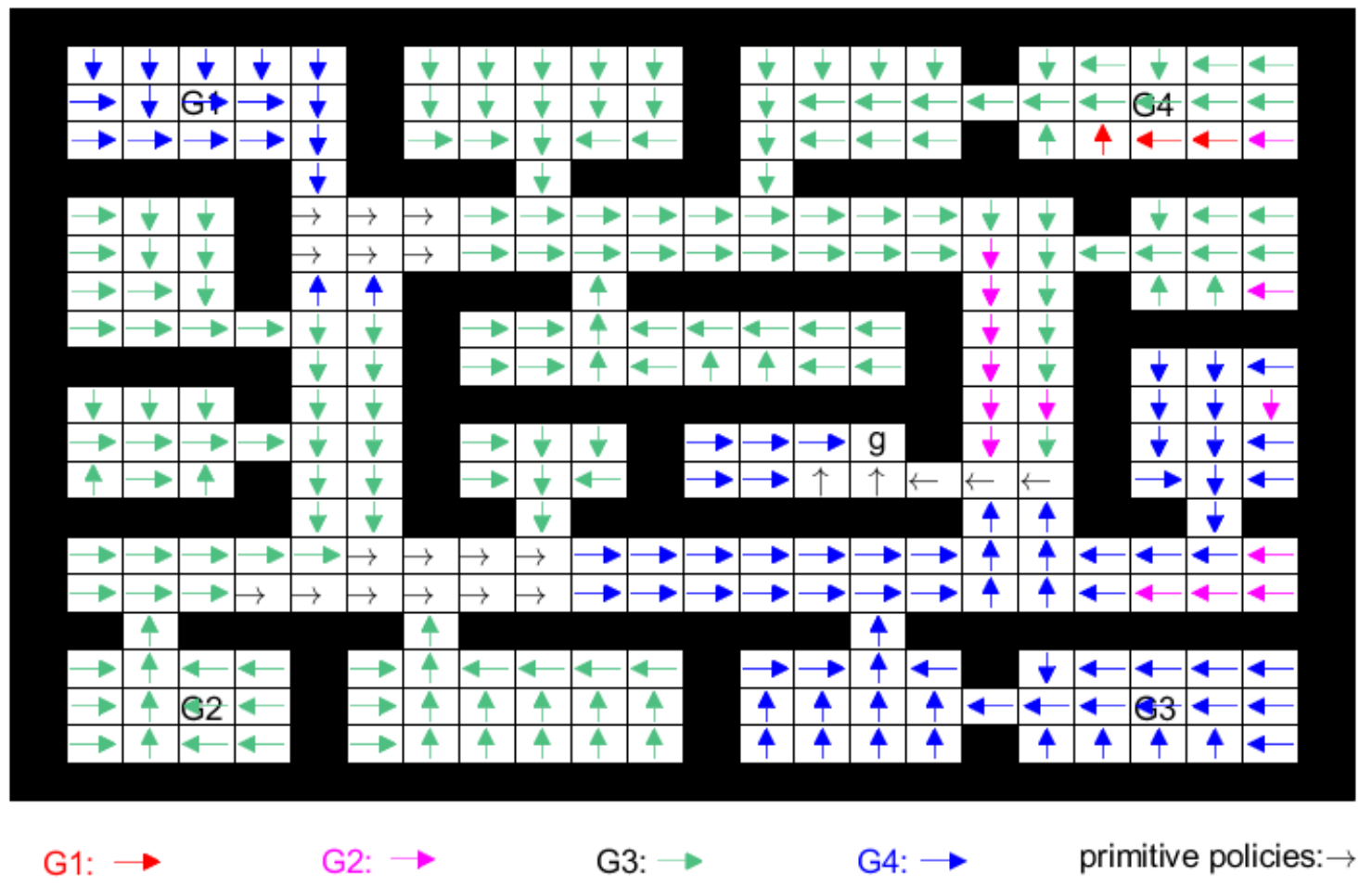}}
	\setlength{\abovecaptionskip}{-8pt}
	\setlength{\belowcaptionskip}{-15pt}
	\caption{Options selected by CAPS to solve task $g$.}
\end{figure}
To better understand this significant outperformance, we illustrate how CAPS reuses source policies in target task $g$. From Figure 3, we can see that CAPS learns to choose the optimal policy to reuse for different contexts.
The colors of the arrows represent the options CAPS takes in a greedy strategy.
The directions of arrows denote the policies of selected options.
For states around goal $g$, no source policy is useful for the target task, so CAPS chooses primitive options. For other states,
CAPS chooses different source options.
For example, source policy $\pi_{G3}$ selected by CAPS navigates an agent out of the room in the lower left corner. We can say that, in some sense, CAPS decomposes a target task to subtasks according to existing knowledge.

\begin{figure*}[hbpt]
	\centering
	\subfigure[source $G1$]{
		\begin{minipage}[b]{0.23\textwidth}
			\includegraphics[width=1\textwidth]{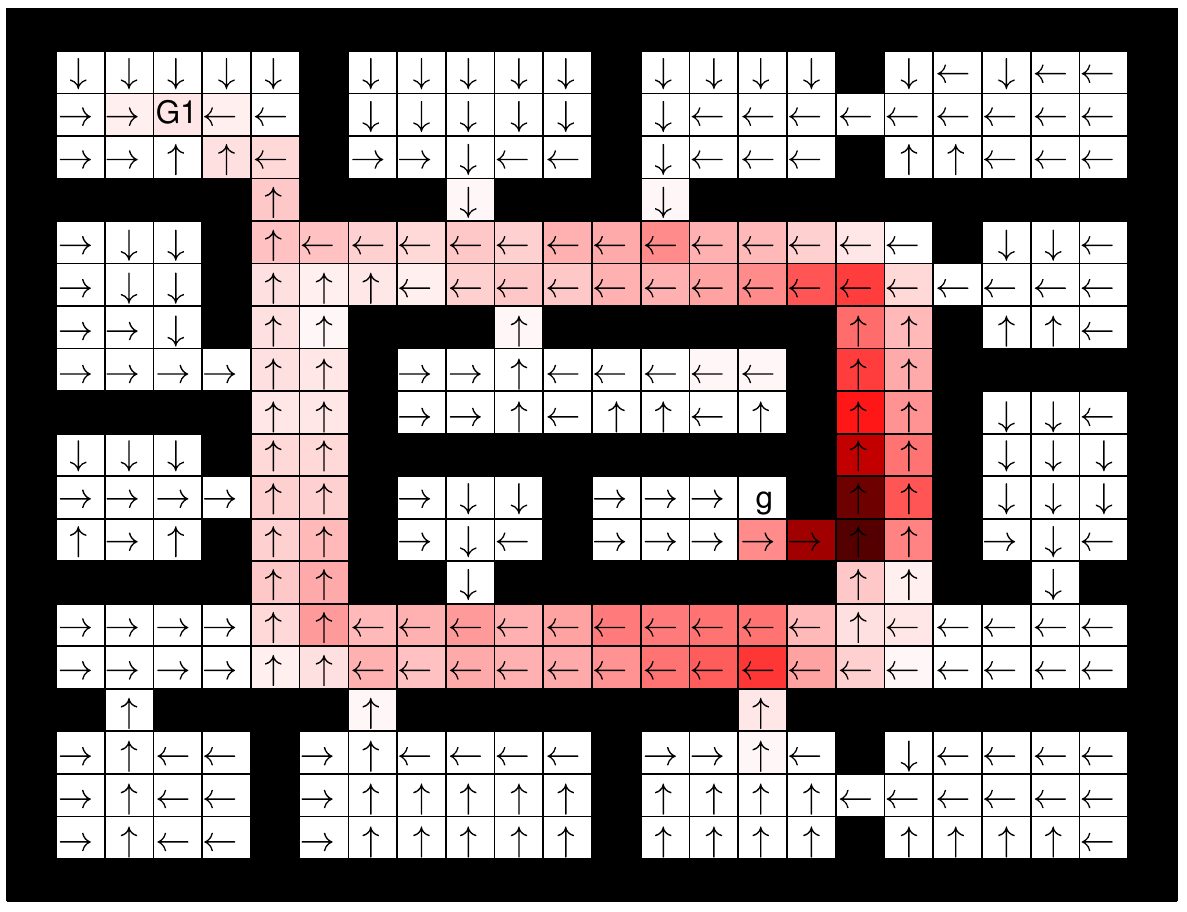}
		\end{minipage}
	}
	\subfigure[source $G2$]{
		\begin{minipage}[b]{0.23\textwidth}
			\includegraphics[width=1\textwidth]{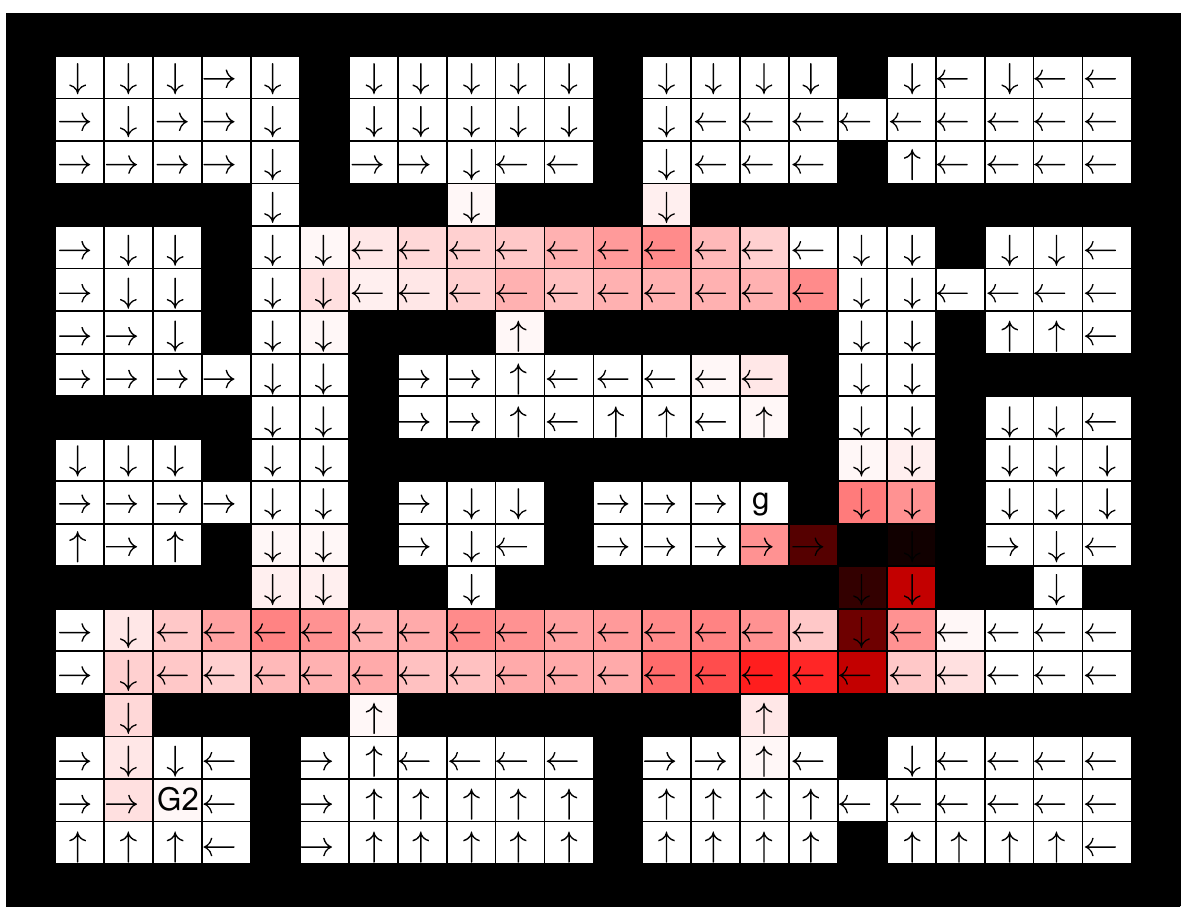}
		\end{minipage}
	}
	\subfigure[source $G3$]{
		\begin{minipage}[b]{0.23\textwidth}
			\includegraphics[width=1\textwidth]{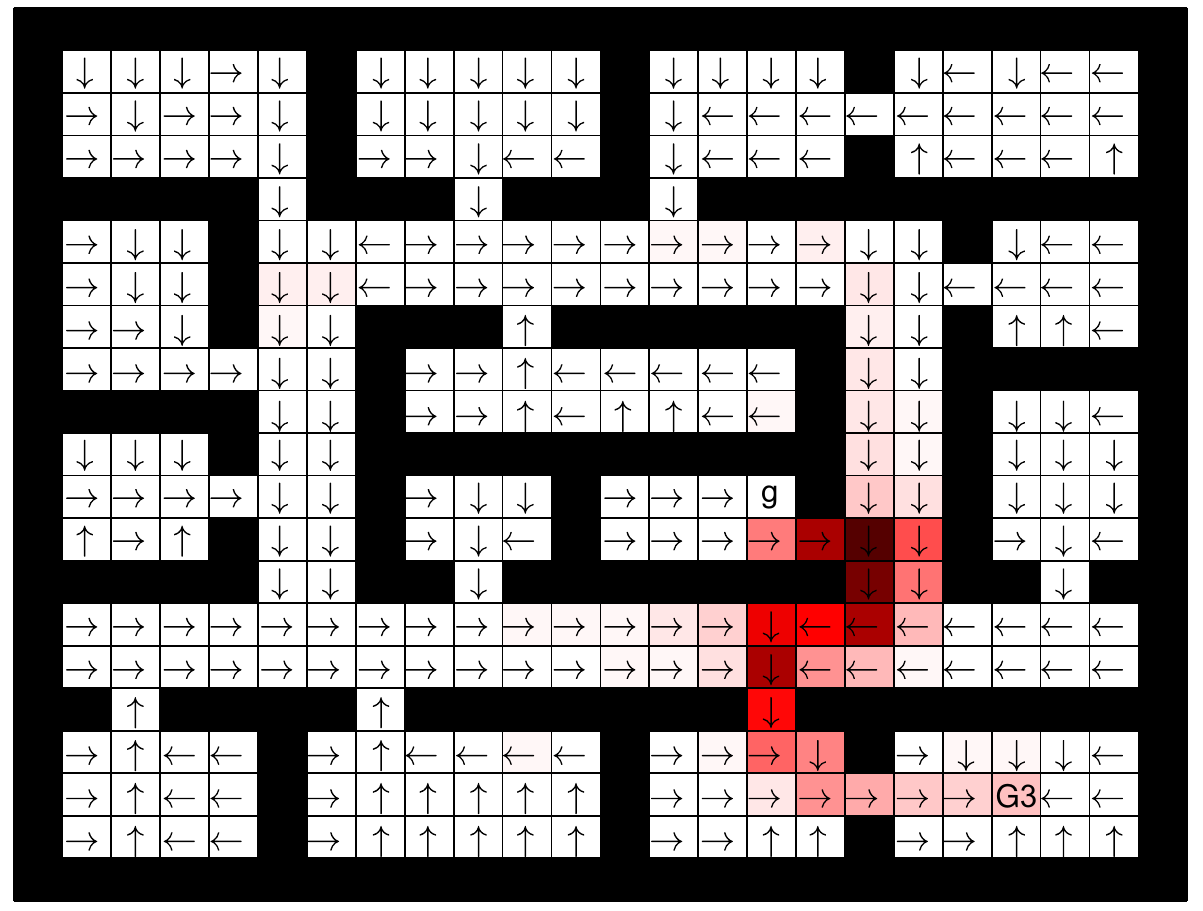}
		\end{minipage}
	}
	\subfigure[source $G4$]{
		\begin{minipage}[b]{0.23\textwidth}
			\includegraphics[width=1\textwidth]{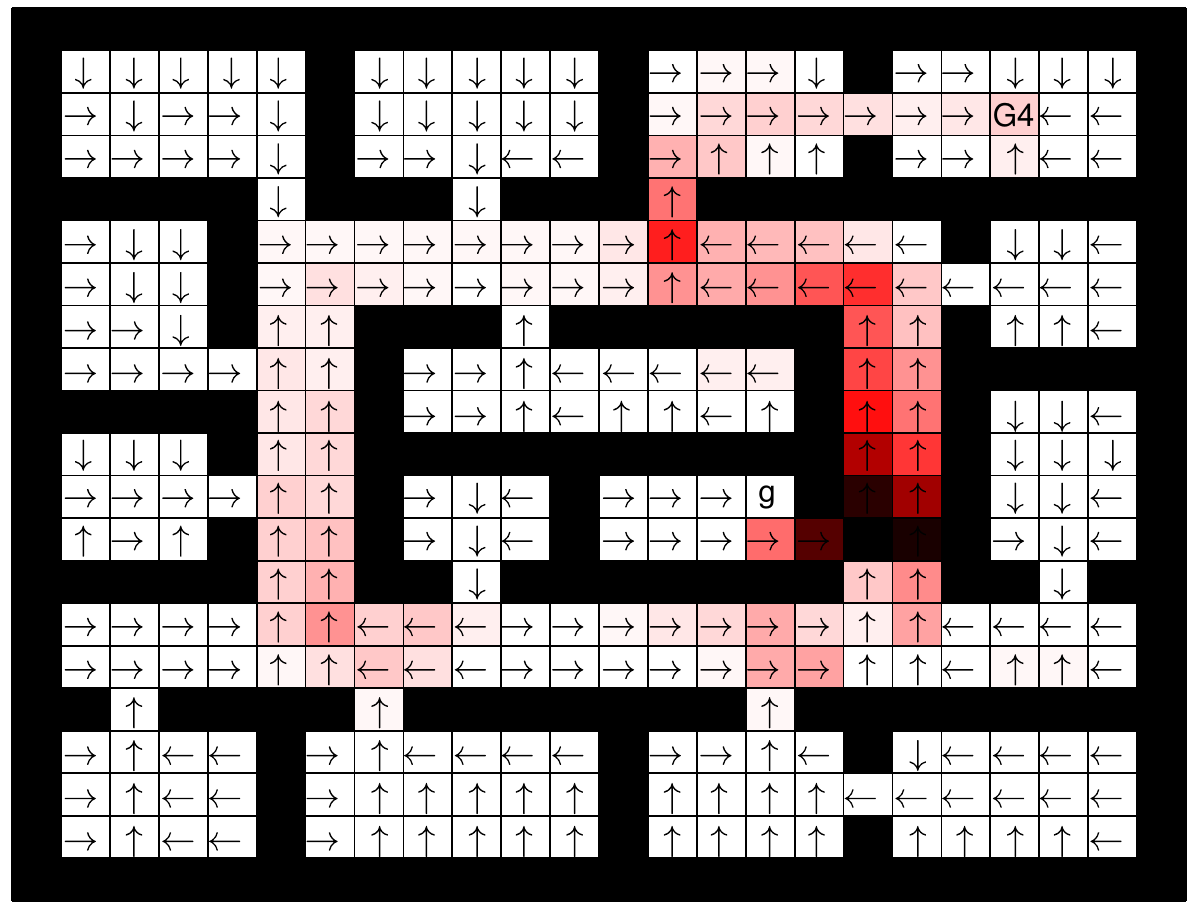}
		\end{minipage}
	}
\setlength{\abovecaptionskip}{-2pt}
\setlength{\belowcaptionskip}{-10pt}
	\caption{Termination probabilities of source policies when solving task $g$. }
	\label{term1}
\end{figure*}

Figure \ref{term1} illustrates the learned termination function for each source policy.
The darker colors encode higher termination probabilities.
The arrows denote the actions taken by each source policy in different states.
From Figure 4, we can observe that the learned termination functions are intuitively sensible.
Large blocks of states with light colors validate that CAPS supports temporally-extended policy reuse.
For states where a source policy is not consistent with the optimal target policy $\pi_g^*$, its reuse is terminated with a high probability, illustrated by a dark red color.

\vspace{-10pt}
\begin{figure}[H]
	\centering
	{
		\includegraphics[width=0.47\textwidth]{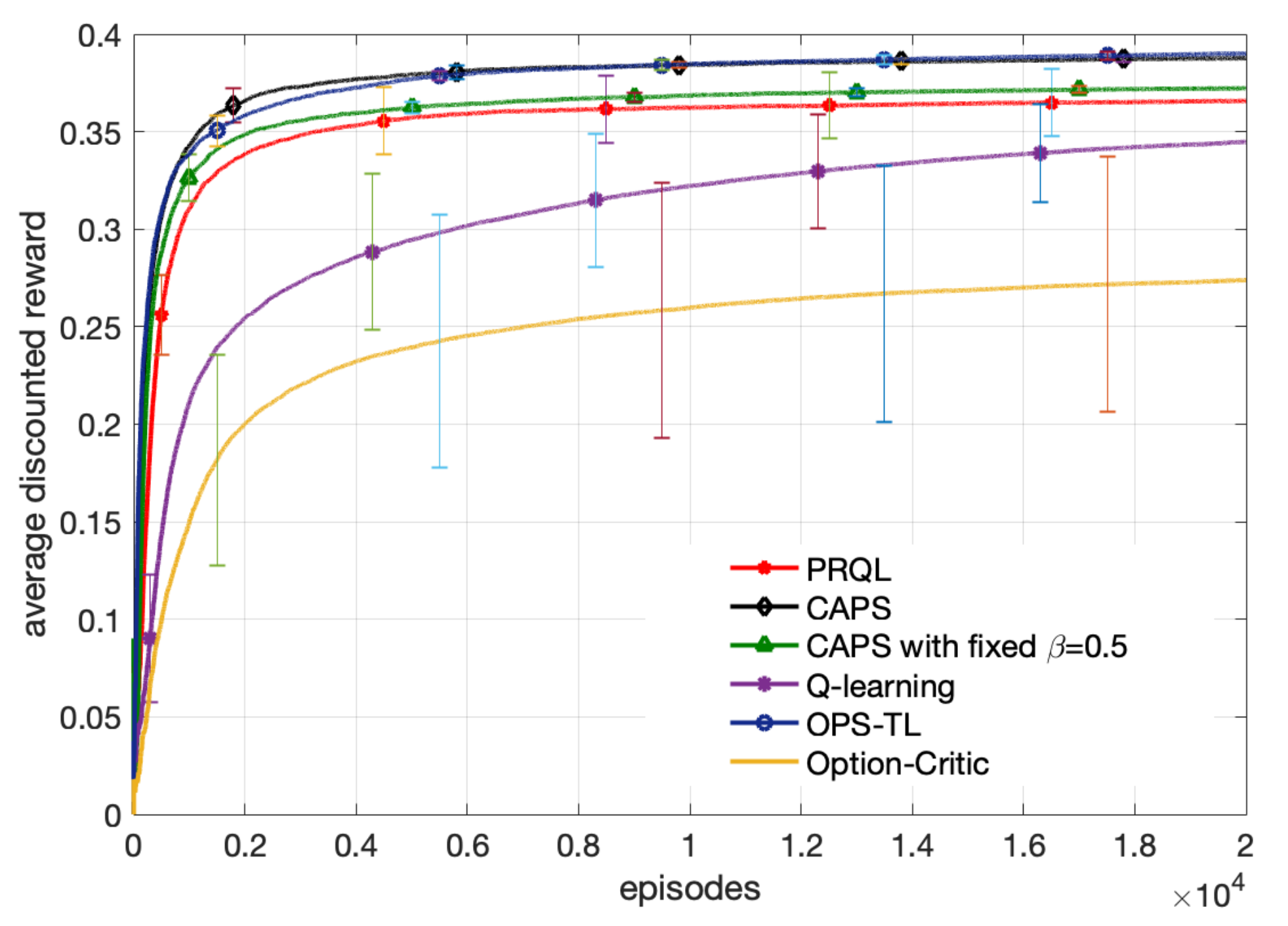}}
	\setlength{\abovecaptionskip}{-10pt}
	\setlength{\belowcaptionskip}{-15pt}
	\caption{Average discounted rewards of CAPS, PRQL, OPS-TL, OC and Q-learning on target task $g'$. }
\end{figure}

Figure 5 shows the learning curves for target task $g'$, where there is a single best policy
$\pi_{G1}$ for reuse. Both CAPS and OPS-TL provably select $\pi_{G1}$. CAPS still performs slightly better than OPS-TL, because OPS-TL's ad hoc hyperparameter for specifying the termination function is hard to tune in practice while CAPS automatically learns termination functions during policy reuse.
This slight outperformance indicates that concurrently learning to identify transfer contexts and selecting the best source policy does not sacrifice the learning performance.
{{OC performs better in task $g'$ than in task $g$, because the adaptation for $\pi_{G1}$ in task $g'$ is much less  and the action space near goal $g'$ in OC is complete.}}

To verify CAPS works as well in situations where transitions between source and target tasks are different, we conduct experiment on target task in Figure 6, whose map is much different from  the map of sources.
The results in Figure 7 shows that CAPS outperforms other methods even if only some parts of source policies can be reused. CAPS identifies the useful parts based on contexts automatically.
\vspace{-5pt}
\begin{figure}[H]
	\centering
	\label{change}
	\begin{minipage}[t]{0.47\textwidth}
		\centering
	{	\includegraphics[width=0.9\textwidth]{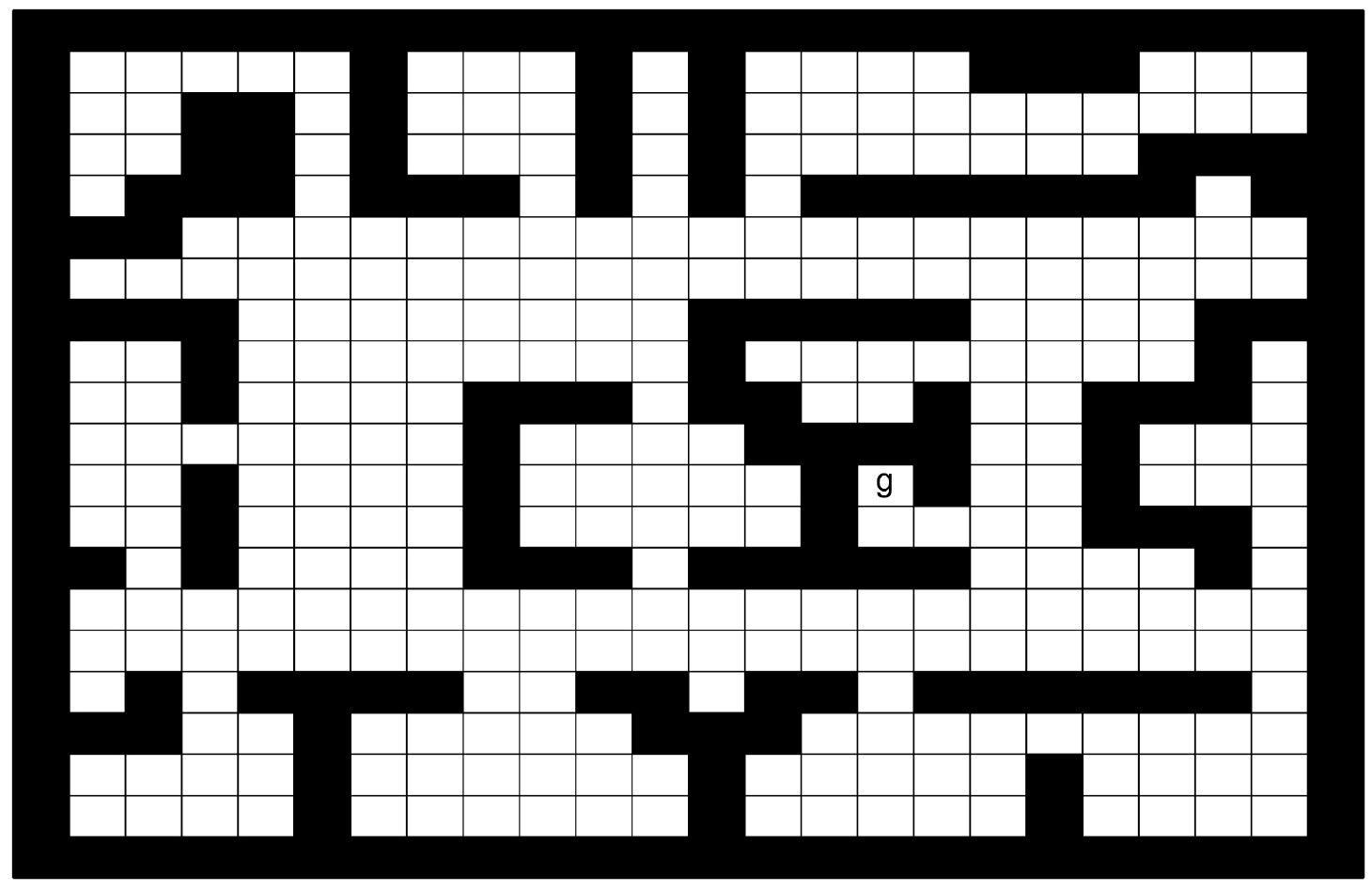}}
	\caption{Target task  of different transitions from source tasks. }
	\end{minipage}
\end{figure}
\vspace{-10pt}
\begin{figure}[H]
\label{reward}
	\begin{minipage}[t]{0.47\textwidth}
	\centering
	{	\includegraphics[width=0.9\textwidth]{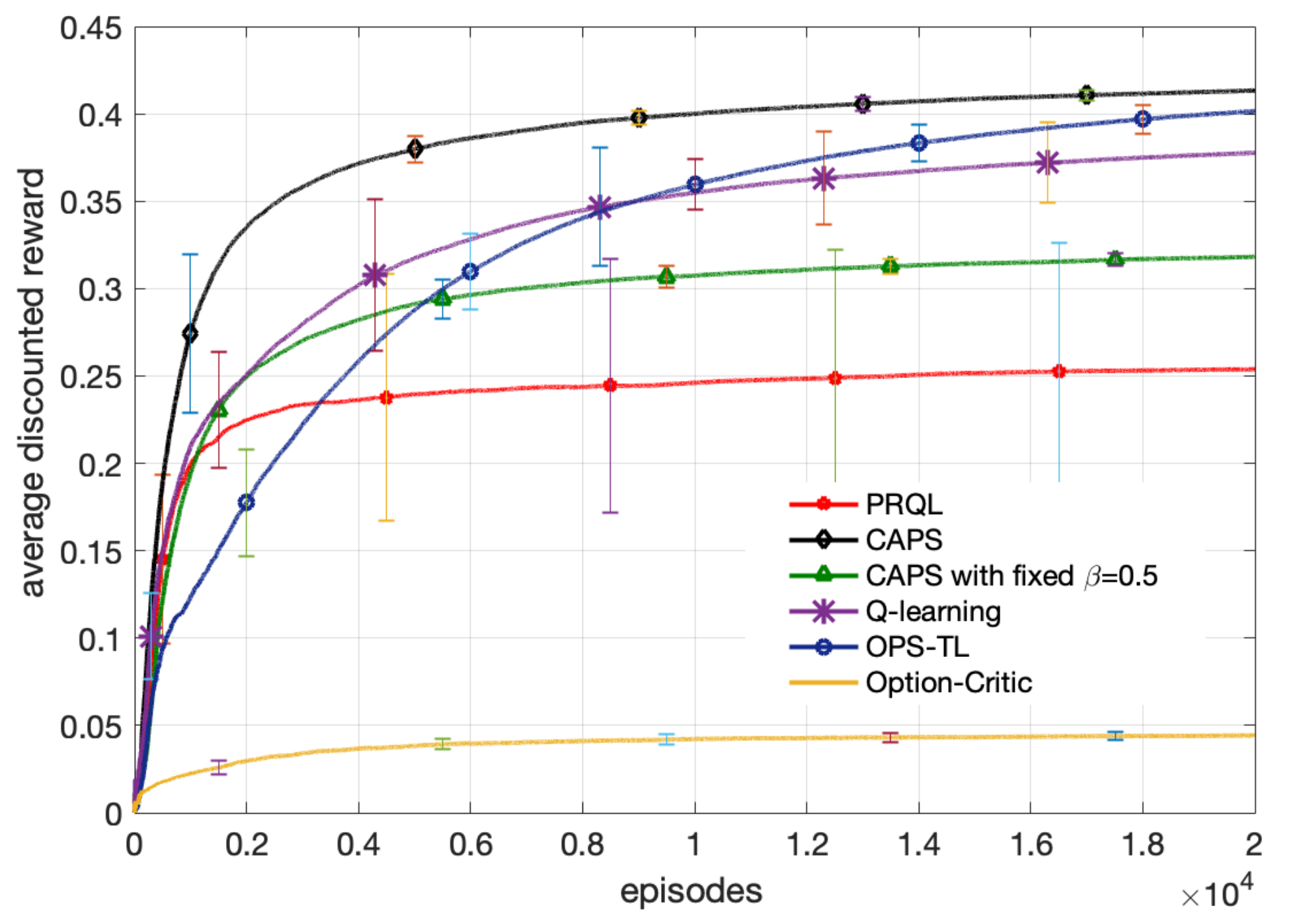}}
	\caption{Average discounted rewards of CAPS, PRQL, OPS-TL, OC and Q-learning on the target task in Figure 6.}
\end{minipage}
\end{figure}
\subsection{Pygame Learning Environment}
\subsubsection{Neural Network Structure}
	\mbox{ }\par
\begin{figure}[hbpt]
	\includegraphics[width = .47\textwidth]{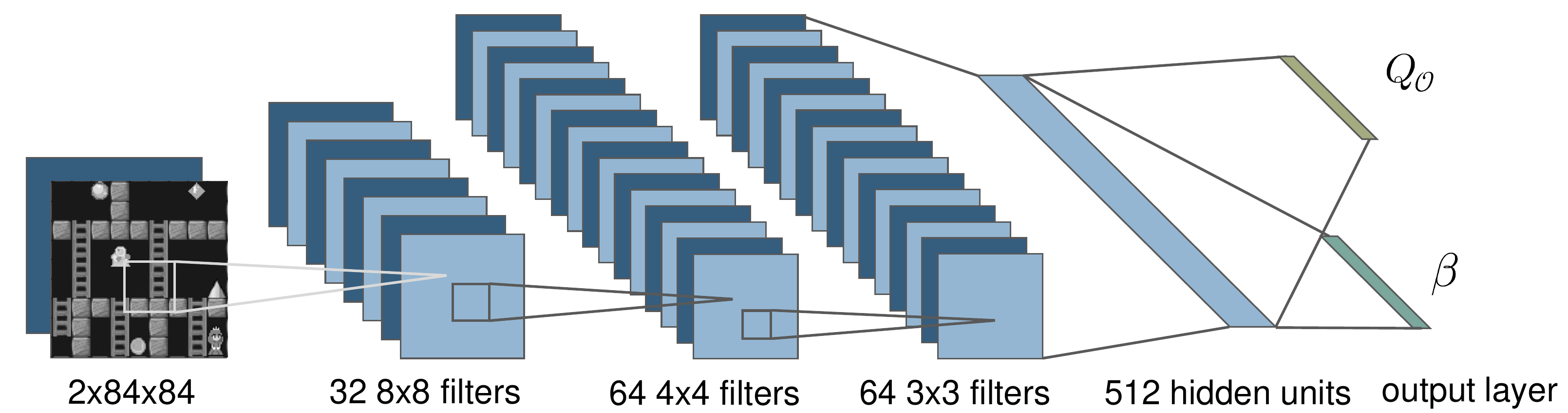}
	\caption{Neural Network Structure.}
	\label{522}
\end{figure}

CAPS is also applicable with a function approximation.
We use a deep neural network to approximate option-value function $Q_{\mathcal O}$ and termination function $\beta$.
Our network structure has the same convolutional structure as DQN \cite{mnih2015human}. There are 3 convolutional layers followed by 2 fully-connected layers shown in Figure \ref{522}.

$ Q_O $ is trained off-policy with experience replay and target network, while $ \beta $  is trained online with fixed learning rate as $0.00025$.
We assume the output of the last but one layer as the learned representations of states,
so we only train the last layer
when learning $\beta$.
We also employ double Q network \cite{van2016deep} and gradient clipping \cite{bengio2013advances}.	

We perform a training step on $Q_{\mathcal O}$ each step with minibatches of size 32 randomly sampled from a replay buffer of one million transitions every 4 transitions encoded into the replay buffer.
The learning rate of $Q_{\mathcal O}$ is annealed piecewise linearly from $10^{-4}$  to $5\times10^{-5}$ over the first $2.5$ million training steps, then fixed at $5\times10^{-5}$ after that.
The training process of $Q_{\mathcal O}$ and $\beta$ begins after $5\times10^4$ transitions.
$\epsilon$  is annealed piecewise linearly from $1$ to $0.05$ over the first $4.375$ million training steps. $\gamma$ is set \mbox{as $0.99$}.
We add a regularization $\rho = 0.005$ to the advantage function in the update function (\ref{update_function}) analogously to \cite{bacon2017option}.

\subsubsection{Experimental Settings}
	\mbox{ }\par
\begin{figure}[htbp]
	\centering
	\subfigure[ ]
	{
		\begin{minipage}[b]{0.2\textwidth}
			{	\includegraphics[width=1\textwidth]{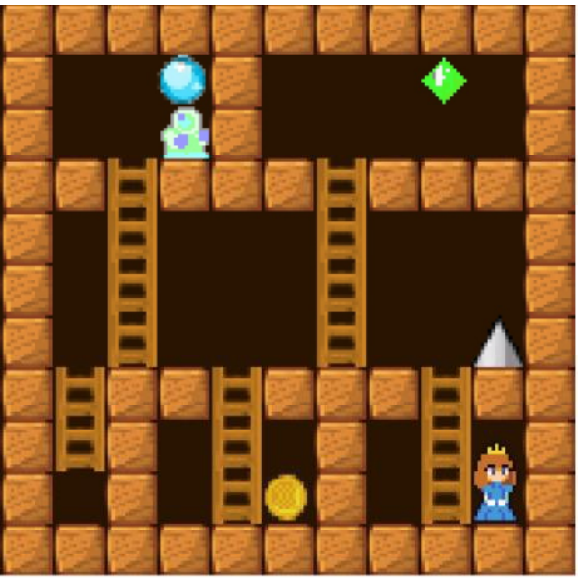}}
			\label{deepg1}
		\end{minipage}
	}
	\subfigure[ ]	{	
		\begin{minipage}[b]{0.2\textwidth}
			{	\includegraphics[width=1\textwidth]{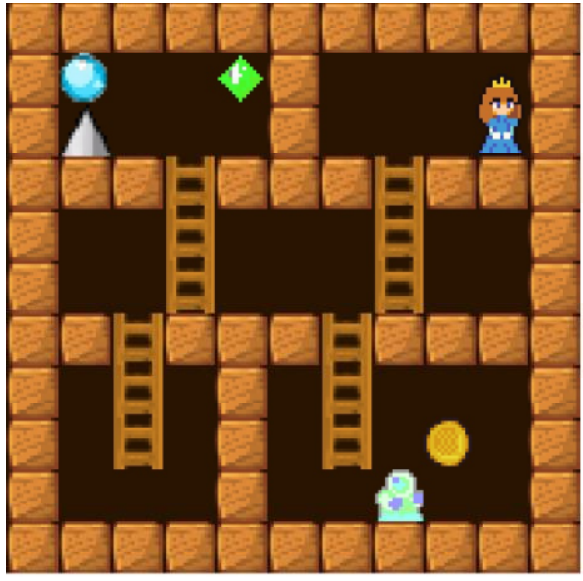}}
			\label{deepg2}
		\end{minipage}
	}
\setlength{\abovecaptionskip}{-2pt}
\setlength{\belowcaptionskip}{-10pt}
	\caption{Two different experimental settings. (a) The goals of source and target tasks are all in different rooms. (b) The goals of one source task and the target task
		are in the same room.}
	\label{deep map}
\end{figure}

Monsterkong of PLE \cite{tasfi2016PLE} is a complex navigation problem with simulated gravity. Two experimental settings are shown in Figure \ref{deep map}.
The character under the blue gem in Figure \ref{deepg1} is an agent, whose initial position is randomly set on bricks.
If the agent reaches a goal, it receives  a reward of 1.
Otherwise, it receives no reward.
The action space
consists of {\em up, down, left, right, jump} and {\em no-op}, six actions.
The agent can move up and down only when it is on a ladder.
Ineffective actions are treated as {\em no-op}s.
An episode terminates
in three cases: the agent reaches the goal, the agent touches a triangle spike or the timesteps exceed horizon $ H=1200 $.

Since the bricks surrounding the images in Figure \ref{deep map} are useless, we clip the bricks and down-sample the remaining part to $84\times84$.
Then we convert the preprocessed images to gray-scale, stack the last two and feed them to the network.

To illustrate the robustness of CAPS, we choose different objects as goals for source and target tasks in the two settings.
In Figure \ref{deepg1}, the green diamond, the blue gem, and the yellow coin are goals for source tasks. The princess is the goal for target task $g1$.
As for task $g1$, there is no explicitly similar source task.
In Figure \ref{deepg2}, the green diamond, the yellow coin, and the princess are goals for source tasks. The blue gem is the goal for target task $g2$, which is in the same room with the green diamond. So there is one remarkably  similar source task in the library to target task $g2$.
\vspace{-8pt}

\subsubsection{Results}
	\mbox{ }\par
\begin{figure*}[htbp]
	\centering
	\subfigure[]{
		\begin{minipage}[b]{0.47\textwidth}
			\label{deep diff}
			{	\includegraphics[width=1\textwidth]{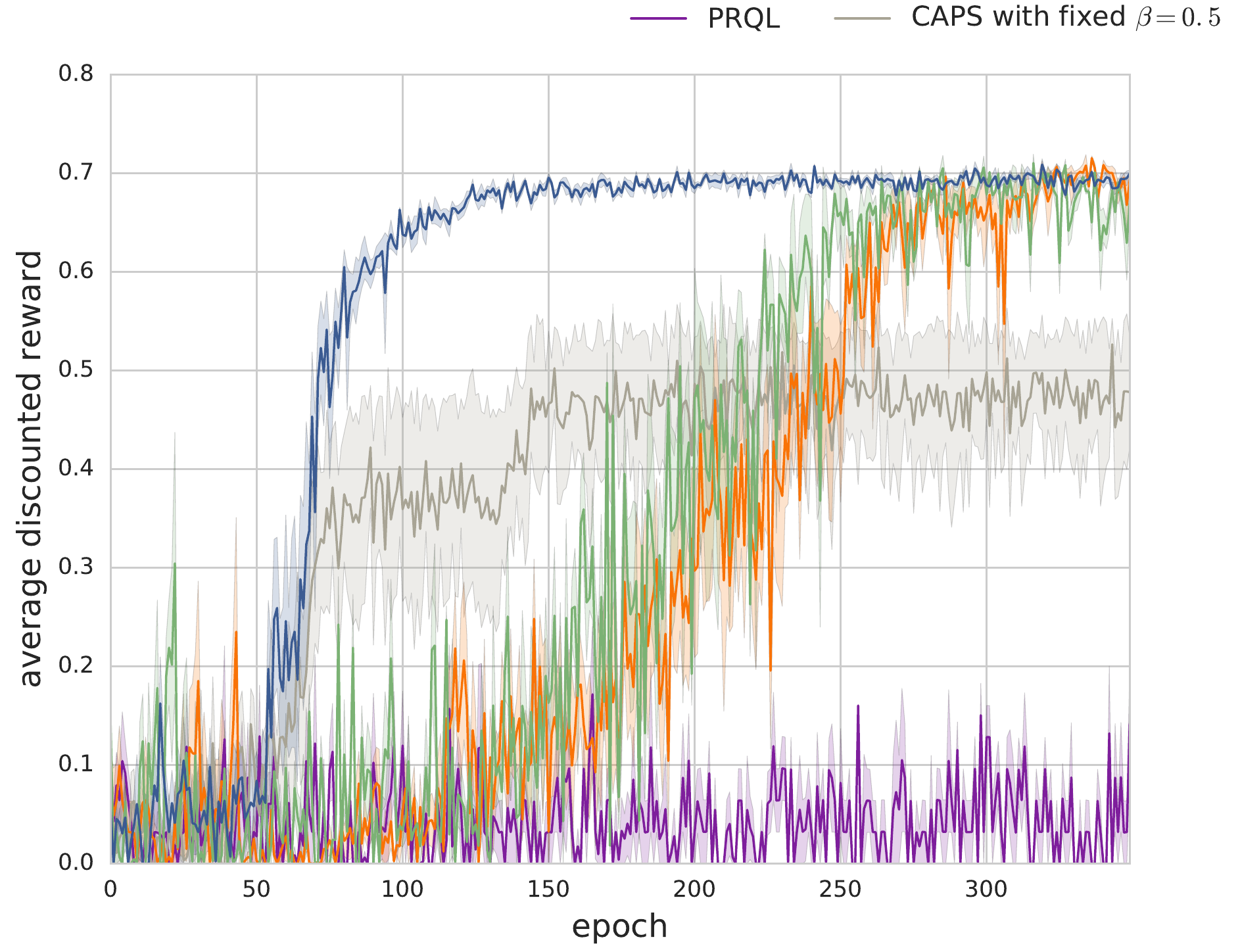}}
		\end{minipage}
	}
	\subfigure[]{
		\begin{minipage}[b]{0.46\textwidth}
			\label{deep same}
			{
				\includegraphics[width=1\textwidth]{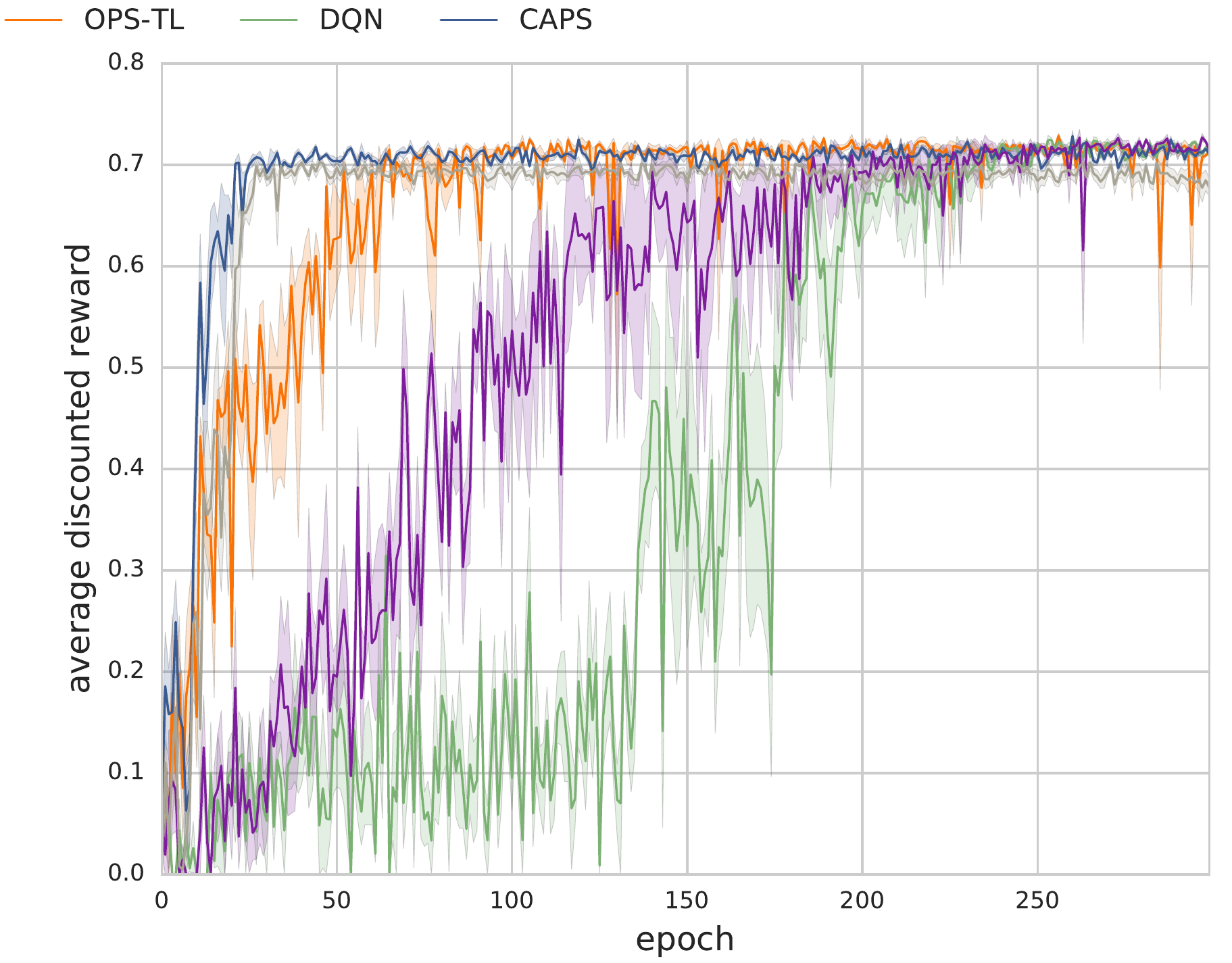}}
		\end{minipage}
	}
\setlength{\abovecaptionskip}{-2pt}
\setlength{\belowcaptionskip}{-10pt}
	\caption{Average discounted rewards of CAPS, PRQL, OPS-TL and DQN on target tasks $g1$ (a) and $g2$ (b) for $5000$-step evaluation per epoch. }
\end{figure*}

Shown in Figure \ref{deep diff}, the average rewards of solving task $g1$ are evaluated 5000 steps every 12500 training steps (one epoch) for CAPS and other baseline methods
The hyperparameters of all methods are tuned to give the best performance for this experiment. Each learning process has been executed for 5 times.

The learning curve of CAPS starts to rise at about 50 epochs and converges to the optimal value in 125 epochs, which is significantly faster than other methods.
Previous methods can only benefit from one source task, so they perform poorly when source tasks
are  much different from the target task. PRQL even  suffers from negative transfers.
With fixed $ \beta=0.5 $, CAPS converges to a suboptimal policy,
which illustrates the importance of a proper termination to the policies selected.
Since the rewards are averaged for only one time evaluation of 5000 steps, the reward curves in this experiment shake more severely than those of last experiment, which are averaged from the start.

Moreover,  we show the performance comparison of solving task $g2$ in Figure \ref{deep same}. CAPS converges to the optimal policy the fastest when there is a source task remarkably  similar to the target task.
Since the source knowledge in this setting is more useful than that of task $g1$, the learning performance of CAPS is  significantly better.
PRQL and OPS-TL also show positive transfers in this experimental setting.

\begin{figure}[htbp]
	\centering
	\subfigure[]{
		\begin{minipage}[b]{0.2\textwidth}
			{	\includegraphics[width=1\textwidth]{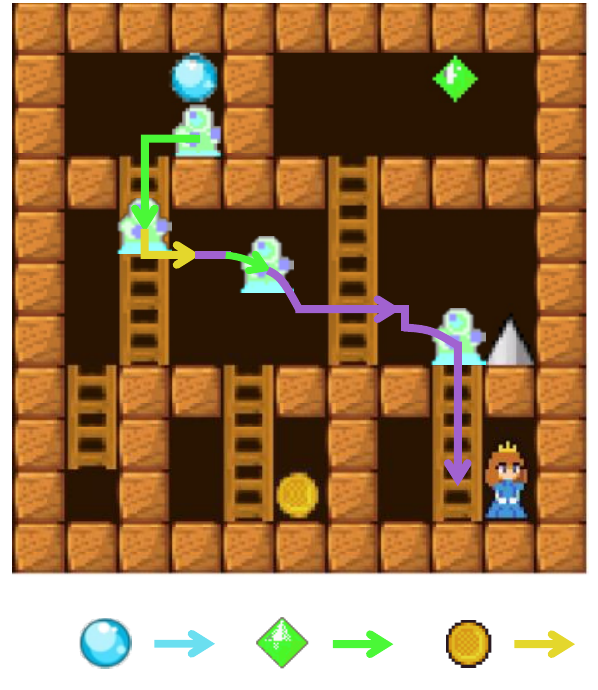}}
			\label{deeppolicyg1}
		\end{minipage}
	}
	\subfigure[]{
		\begin{minipage}[b]{0.2\textwidth}
			{
				\includegraphics[width=1\textwidth]{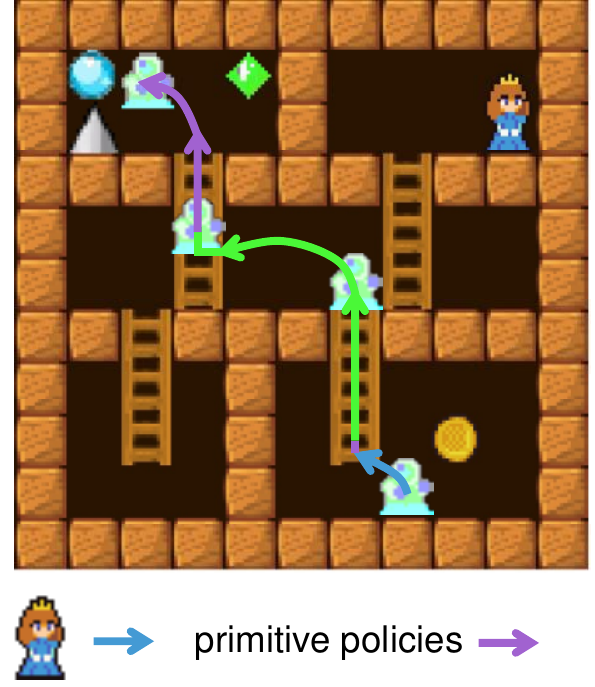}}
			\label{deeppolicyg2}
		\end{minipage}
	}
\setlength{\abovecaptionskip}{-2pt}
\setlength{\belowcaptionskip}{-15pt}
	\caption{Trajectories of the agent for an episode
		to solve task $g1$ (a) and $g2$ (b).}
	\label{deep policy}
\end{figure}

We further demonstrate how CAPS choose policies to reuse in Figure \ref{deep policy}. The arrows in each figure show a complete trajectory of the agent from an initial position to the goal.
The colors of arrows denote different policies the agent selects.
In Figure \ref{deeppolicyg1},
at the beginning, the agent chooses a source policy with the green diamond as its goal to navigate out of the room.
After that, the agent switches to another source policy to get closer to the princess.
Finally, since goals of source and target tasks are all  in different rooms, the agent has to utilize primitive policies to reach the goal of task $g1$.
\vspace{-12pt}

\section{Summary and Discussions}
In this paper, we develop a multi-policy reuse method, called {\em Context-Aware Policy reuSe} (CAPS), that leverages knowledge from multiple source policies and greatly accelerates reinforcement learning. 
Unlike previous works on top-policy learning and policy reuse, CAPS not only optimally learns when and which source policy to reuse, but also when to terminate its reuse to support temporally-extended policy reuse. 
In addition, CAPS provides the same optimality guarantee of the target policy learning as Q-learning, assuming no prior knowledge about the models of the target task and source tasks.
CAPS versus Q-learning is like A* versus best-first search, providing a mechanism for speeding up the learning while keeping the optimality guarantee.
Results from both toy experiments and deep-learning experiments show that CAPS significantly outperforms other state-of-the-art policy reuse methods, and verify that effectively and concurrently utilizing multiple source policies is crucial to improve transfer efficiency.

In our experiments, although the size of the augmented policy library  is larger than that of the action space in the original problem, CAPS still significantly outperforms Q-learning. One reason is that, although depending on the quality of source policies, the probability of selecting useful options can be larger than selecting an optimal action on many states. Another reason is that CAPS supports temporally-extended policy reuse and do not need to make a decision of choosing a policy at each step. It is possible for CAPS to underuse source policies when the dimension of the action space is much larger than the number of source policies. In such situations, we can employ up-sampling techniques to improve the probability of reusing source policies instead of selecting primitive policies.

One advantage of CAPS is that it assumes no constraints on the representation of the source policies and no prior knowledge about goals and transition functions of the source and target tasks, which is different from the approach of universal value function approximator \cite{schaul2015universal}. When there is some prior knowledge about which source policy is better to reuse, we can use it to shape the exploration of CAPS to speed up the learning. To support lifelong learning, it is important to identify whether a new policy is necessary to be added to the policy library, which is part of future work. To further improve the reusability of source policies, we will also explore to learn the initiation state sets for options as well as their termination functions.

\begin{acks}
	
	The authors would like to thank the anonymous reviewers for their valuable comments and helpful suggestions. The work is supported by Huawei Noah‘s Ark Lab under Grant No. YBN2018055043.

\end{acks}

\appendix
\section{Implementation Details  of Option-Critic}
The source policies suitable to the downstream tasks in the Option-Critic (OC) framework need to be differentiable. In contrast, our method has no requirement for the representation of the source policies. So we first  train the source tasks using the OC model with only one intra-option policy for $1e5$ episodes until convergence, so the intra-option policies can be regarded as source policies in our setting. Then we train  a new OC model to learn a target task. 
The intra-option policy and inter-option policy are softmax policies and the termination functions are sigmoid functions.
The new OC model has 4 intra-option policies, which are initialized with the intra-option policies learned in the source tasks. The inter-option policies and termination functions are learned from scratch.
 In target task, the low-level and high-level policies are simultaneously trained, the same as what OC did in their original paper.
The learning rate for all the policies in OC is set as $0.05$ and the temperature for the softmax policy is $1e-5$.